\documentclass[letterpaper, 10 pt, journal, twoside]{IEEEtran}
\usepackage{cite}
\usepackage{amsmath,amssymb,amsfonts}
\usepackage{algorithmic}
\usepackage{graphicx}
\usepackage{textcomp}
\usepackage{blindtext}
\usepackage{url}
\usepackage{hyperref}
\usepackage[pscoord]{eso-pic}

\newcommand{\placetextbox}[3]{
  \setbox0=\hbox{#3}
  \AddToShipoutPictureFG*{
    \put(\LenToUnit{#1\paperwidth},\LenToUnit{#2\paperheight}){\vtop{{\null}\makebox[0pt][c]{#3}}}%
  }%
}%

\begin{document}

\placetextbox{0.5}{1}{This is the author's version of an article that has been published in IEEE Transactions on Industrial Informatics.}
\placetextbox{0.5}{0.97}{The final version of record is available at \href{https://doi.org/10.1109/TII.2024.3393140}{https://doi.org/10.1109/TII.2024.3393140}}%
\placetextbox{0.5}{0.05}{Copyright \copyright 2024 IEEE. Personal use is permitted, but republication/redistribution requires IEEE permission.}

\title{C(NN)FD - a deep learning framework for turbomachinery CFD analysis}

\author{Giuseppe Bruni,  
       Sepehr Maleki,
       Senthil K. Krishnababu 
       \thanks{This work was supported and funded by Siemens Energy Industrial Turbomachinery Ltd.}
       \thanks{Giuseppe Bruni, Sepehr Maleki and Senthil K. Krishnababu are with the University of Lincoln, United Kingdom}
       \thanks{Giuseppe Bruni (e-mail: giuseppe.bruni(at)siemens-energy.com) and Senthil K. Krishnababu are with Siemens Energy Industrial Turbomachinery Ltd. }}

\maketitle

\begin{abstract}
Deep Learning methods have seen a wide range of successful applications across different industries. Up until now, applications to physical simulations such as CFD (Computational Fluid Dynamics), have been limited to simple test-cases of minor industrial relevance. This paper demonstrates the development of a novel deep learning framework for real-time predictions of the impact of manufacturing and build variations on the flow field and overall performance of axial compressors in gas turbines, with a focus on tip clearance variations. The scatter in compressor efficiency associated with manufacturing and build variations can significantly increase the $CO_2$ emissions, thus being of great industrial and environmental relevance. The proposed \textit{C(NN)FD} architecture achieves in real-time accuracy comparable to the CFD benchmark. Predicting the flow field and using it to calculate the corresponding overall performance renders the methodology generalisable, while filtering only relevant parts of the CFD solution makes the methodology scalable to industrial applications.
\end{abstract}


\section{Introduction}

Gas turbine manufacturers have accumulated a wealth of operational data over the past decades, which has allowed establishing best practices and design guidelines for their manufacturing and build processes. The tolerance ranges are often defined based on a trade-off between cost and engine performance. However, the estimated impact on performance used to define admissible ranges, is often based on previous experience and simplified correlations. In parallel, over the years the scientific community has addressed many of the fundamental challenges affecting the compressor aerodynamic performance. Significant advancements have been made in understanding the aerodynamic phenomena relating to the compressor operation and the associated modelling strategies required to characterize it. The accuracy of CFD analyses, has made them an integral part of the industrial design process, often in conjunction with optimization strategies. The design capabilities of gas turbine manufacturers have therefore greatly improved over the past decades, closely approaching the theoretical limit for the current technology. This advancement has led to an interest in the development of robust design strategies to account for real-world effects, such as manufacturing and build variations, in the design and analyses processes. Current research is mostly focused on the modelling and understanding of the features driving performance variations. However, there is a further requirement to translate that understanding into usable low-order models with appropriate predictive capabilities. The focus of this work is therefore to demonstrate the efficacy of the proposed deep learning framework to model the effect of manufacturing and build variations.

\textbf{The challenge}: Manufacturing and build variations are known to significantly affect the performance of gas turbines. One of the main contributors to the engine performance and operability is the axial compressor. The aerodynamic performance of axial compressors is traditionally predicted using CFD. These calculations are performed only for an idealized "as designed" geometry, which is known to be different from the specific build of each engine. For a given compressor design, the overall efficiency is significantly affected by a combination of in-tolerance geometrical variations, which can significantly increase the $CO_2$ emissions of the gas turbine. Therefore, predicting the impact of manufacturing and build variations on the engine performance, within quick timescales, is of significant industrial and environmental relevance. 

\textbf{Our contribution}: This paper presents a novel deep-learning framework for the real-time prediction of how manufacturing and build variations affect the flow field and the overall performance of the compressor. The proposed framework achieves accuracy close to that of the standard CFD solvers, within a significantly shorter timescale. The proposed C(NN)FD architecture aims to provide a generic framework, which is envisioned to be incorporated as part of the manufacturing and build process, without the associated computational cost and requirement for specialised engineers to carry out CFD analyses.

The effect of build and manufacturing variability on the overall performance and the associated aerodynamic effects have been widely reported in literature \cite{Montomoli2018} \cite{Wang2020}. For axial compressors applications, variations in tip clearance, surface roughness and airfoil geometry are known to significantly affect the overall performance and stability margin. The focus of the current work is on tip clearance variations, as they are one of the main sources of performance variability. However, this framework is readily generalisable to other manufacturing variations, which will be the focus of future research.

\subsection{Tip Clearance variation background}

\begin{figure}[!htbp]
    \centering
       \includegraphics[width=0.45\textwidth]{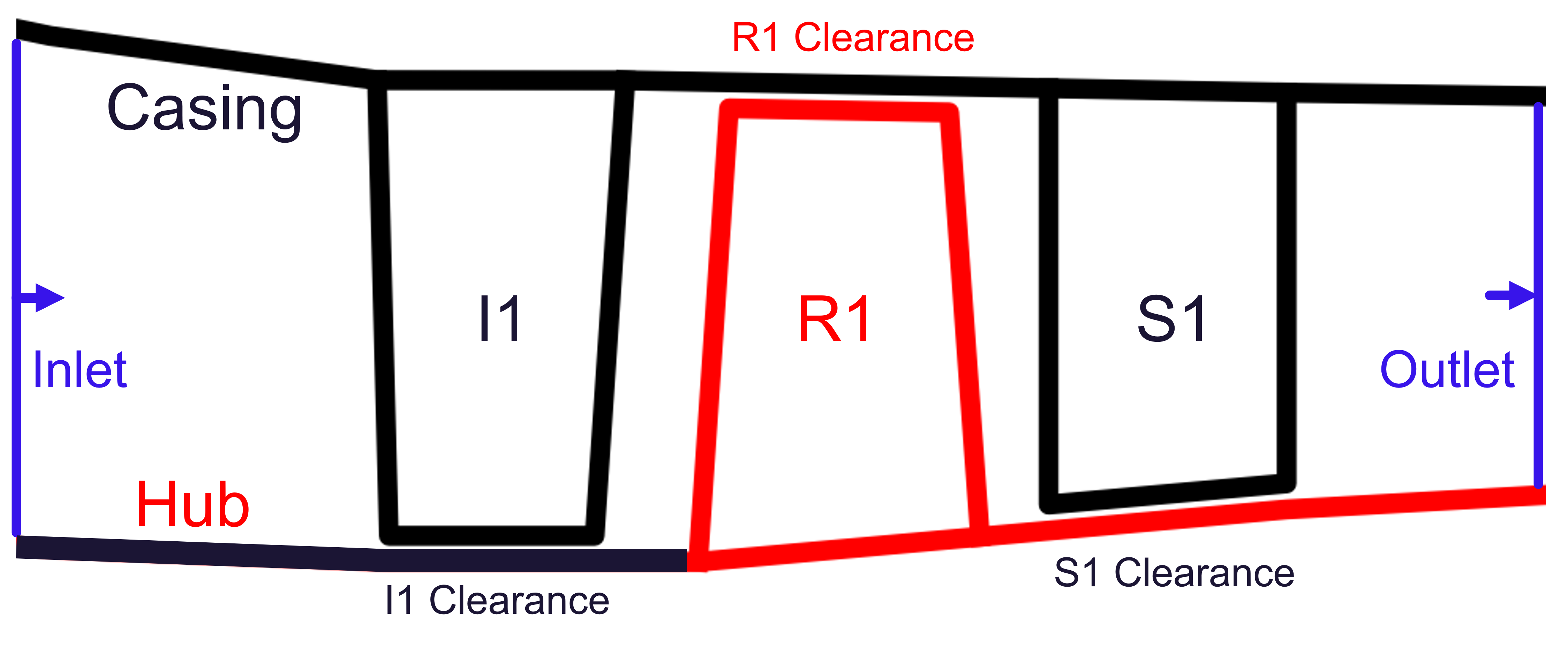}
       \caption{Overview of the 1.5 stage compressor gas path considered}
       \label{fig:Domain_sketch}
\end{figure}

The term tip clearance refers to the radial distance between rotating and stationary components in turbomachinery, such as between rotor blades and casing or between stator blades and rotor hub, as shown in Fig. \ref{fig:Domain_sketch}. The gas path sketched represents the 1.5 axial compressor stage considered in this paper, with rotating surfaces marked in red and stationary surfaces in black. 

Tip clearances should be large enough to limit rubbing in operation, while small enough to avoid excessive aerodynamic losses. Axial compressors are designed to operate at a defined tip clearance, which can vary across the stages of a given compressor. The "cold clearance" is specified during manufacturing and build processes, and can be measured in stationary conditions. This differs from the clearance during operation, which is traditionally referred to as "hot clearance" or "running clearance". Due to thermal expansion and centrifugal effect, the hot clearance can be significantly smaller than the cold clearance. The cold clearance values are specified by a target value and a tolerance range, which is set by gas turbine manufacturers based on experience as a trade-off between excessive rubbing on the casing during operation and minimizing tip leakage losses. The cold clearance can therefore vary within the tolerance range specified, and so does the hot clearance compared to the ideal value assumed in standard CFD models. The aerodynamic mechanisms driving tip leakage flows have been actively researched for several decades \cite{Storer1991} and can now be modelled with different degrees of fidelity. Steady state CFD calculations are traditionally used to model the effect of variations in tip clearance size on the overall performance, defining an exchange rate with the efficiency for a specific compressor design. This has been demonstrated in literature for a range of tip clearances and different operating conditions \cite{Sakulkaew2013}, and can be used to drive design changes to target tip leakage loss reductions \cite{Bruni2019}. Higher fidelity methods such as LES (Large-Eddy-Simulations) \cite{Han2021} or DNS (Direct-Numerical-Simulation) \cite{Maynard2022} can also be used to address known shortcomings of typical steady state CFD. However, their high computational cost limits their practical usage to specific applications. As per standard practice, the effect of tip clearance variations is modelled using steady single passage CFD, instead of full-annulus calculations, thus neglecting circumferential geometrical variations in tip clearance. This has been investigated on NASA Rotor-67 \cite{Suriyanarayanan2022}, using a combination of single-passage, double-passage and full annulus calculations. When considering a range of non-dimensional clearances between 0.376\% span and 0.752\% span, typical of manufacturing variations, it was noted that the results of full annulus, double-passage and single-passage with average tip clearance was matching closely. This implies that the effects of non-axisymmetric variations can be neglected when considering manufacturing variations with reasonably tight tolerance ranges. Moreover, manufacturing experience and historical measurement suggest that no significant circumferential variation is expected during the build process. However, circumferential non-uniformities could potentially be relevant for in-service degradation or for casing distortion, and will be addressed in future work.

\subsection{Machine Learning applications to turbomachinery}

Over the recent years, the application of machine learning methodologies to aerodynamic and aeromechanic predictions of turbomachinery components has been investigated intensively. An overview of the current state of the art has been provided by He \cite{He2022}, with a focus on application to turbulence modelling, manufacturing tolerance variations and flutter modelling. A fully connected neural network architecture was used by Krishnababu et al. \cite{Krishnababu2021} to predict the effect of multi-stage compressor tip clearance variations on the overall compressor efficiency, which was found to have good agreement with CFD predictions and to describe the uncertainty of test data. The focus of the work was on a specific compressor design and for a given operating condition. For a more generalized approach, including more operating conditions and different compressor designs, a more complex architecture and more training data would be required. This was partially addressed with the application of a Physics-Informed machine learning pipeline to forced response predictions of an axial compressor as presented by Bruni et al. \cite{Bruni2022}, including the prediction of intermediate outputs and the use of the relevant physical equations required to calculate the target variables of interest. This approach was found to improve predictions and reduce the amount of training data required, compared to directly predicting the target variable with a single fully connected neural networks. However, the results are still expected to be specific to the compressor design considered and not generally applicable. This could be addressed either by significantly extending the training set, or by developing reduced order models which can further describe the aerodynamic and mechanical behaviour driving the forced response prediction in more detail. The use of output consolidation was also shown to increase the accuracy of the overall performance predictions on Rotor-37 by Pongetti et al. \cite{Pongetti2021}. Introducing intermediate models to predict the blade surface static pressure field and exit flow contours increased accuracy of the predictions, compared to a more direct approach of predicting the target overall performance directly. However, the resulting predictions would not be sufficiently accurate to differentiate the effect of manufacturing deviations. This suggests that in order to be able to predict the overall performance variations with reasonable accuracy and in a generalisable manner, accurate predictions of the flow field are required. Then, these predictions can be averaged and used to calculate the overall performance figure of interest, using the relevant physical equations that describe the problem. Using artificial neural networks as a replacement for CFD solvers is currently an active area of research. A common methodology is using Convolutional Neural Networks (CNNs), to predict the flow variables of interest on a Cartesian grid. An example is a U-Net architecture trained on CFD data, for the predictions of the pressure and velocity distributions around a 2D airfoil \cite{Thuerey2020}. However, this approach is limited to very simple geometries that can be meshed with Cartesian grids. Complex geometries for turbomachinery applications are better described by point-wise formulations, which can use structured multi-block or unstructured meshes as inputs. This has been demonstrated successfully for simple geometries using the PointNet architecture \cite{Kashefi2021}, which was then applied to predict the 3D flow field and overall performance parameters of Rotor-37 \cite{Perrone2022}. The limitation of point-wise methods is that the 3D flow field predictions would need to be based on the same computational grid used for training, which renders the approach inflexible and not scalable to industrial applications, where predictions for different engine designs would be required. Lately, Graph Neural Networks (GNNs) have shown promising results with unstructured meshes and irregular geometries for both steady state \cite{Harsch2021} and unsteady \cite{Pfaff2021} aerodynamic predictions. An application to turbomachinery was presented by Li \cite{Li2022}, demonstrating excellent agreement regarding both flow field and overall performance predictions, even tough only a single row turbine application was considered. 

\subsection{Novel solution to machine learning scalability}
Supervised learning methods to predict the full 3D flow field based on CFD solutions have been demonstrated so far only for small computational domains.  However, the computational cost associated with training and the amount of data required, suggests that the approach would not be scalable to industrial applications, especially for multi-stage compressors applications. Standard computational meshes are in the order of magnitude of 10 million nodes, posing challenges from a machine learning perspective, due to the size of the model required for such an application, and also practicality in deploying such an architecture. Storing the full CFD data for all the training database and future simulations is likely be in the order of magnitude of petabytes for a typical engine manufacturer. One solution to the scalability problem of predicting the whole flow field solution is to define a pre-processing step, in which only the data relevant from an engineering perspective is extracted from the computational domain. This can include: blade-to-blade planes, blade surfaces, axial cuts, loading distributions, radial distributions, stage-wise performance and overall performance. Aerodynamic engineers would typically rely on this data during the design and analysis of turbomachinery for the majority of applications, while close examination of the 3D flow field of a CFD solution are less common and generally limited to non-standard cases. The data so identified can then be interpolated to a simplified grid, with a resolution deemed acceptable for the application of interest. Only selected fundamental variables would be retained, while the derived ones can be computed using the appropriate physical equations. This allows to greatly reduce the computational cost, without any loss in accuracy, as only the relevant data required to calculate the target variables are considered. The regression problem is therefore reduced in dimensionality and complexity, unlocking the potential to use U-Net architectures to predict complex 3D turbomachinery flow fields.

\section{Methodology}

A 1.5-stage domain, representative of a modern industrial axial compressor is considered, consisting of three rows, IGV (Inlet-Guide-Vane), Rotor and Stator, each of which with a given blade count defining the number of blades circumferentially. This domain will be extended in future work to a full compressor domain, which can consists of 10 or more subsequent stages. Moreover, only tip clearance is currently considered as input variables for simplicity. Future work will include other manufacturing and build variations, such as surface roughness and geometry variations.

\subsection{Data Generation}
\begin{figure}[!htbp]
    \centering
       \includegraphics[width=0.45\textwidth]{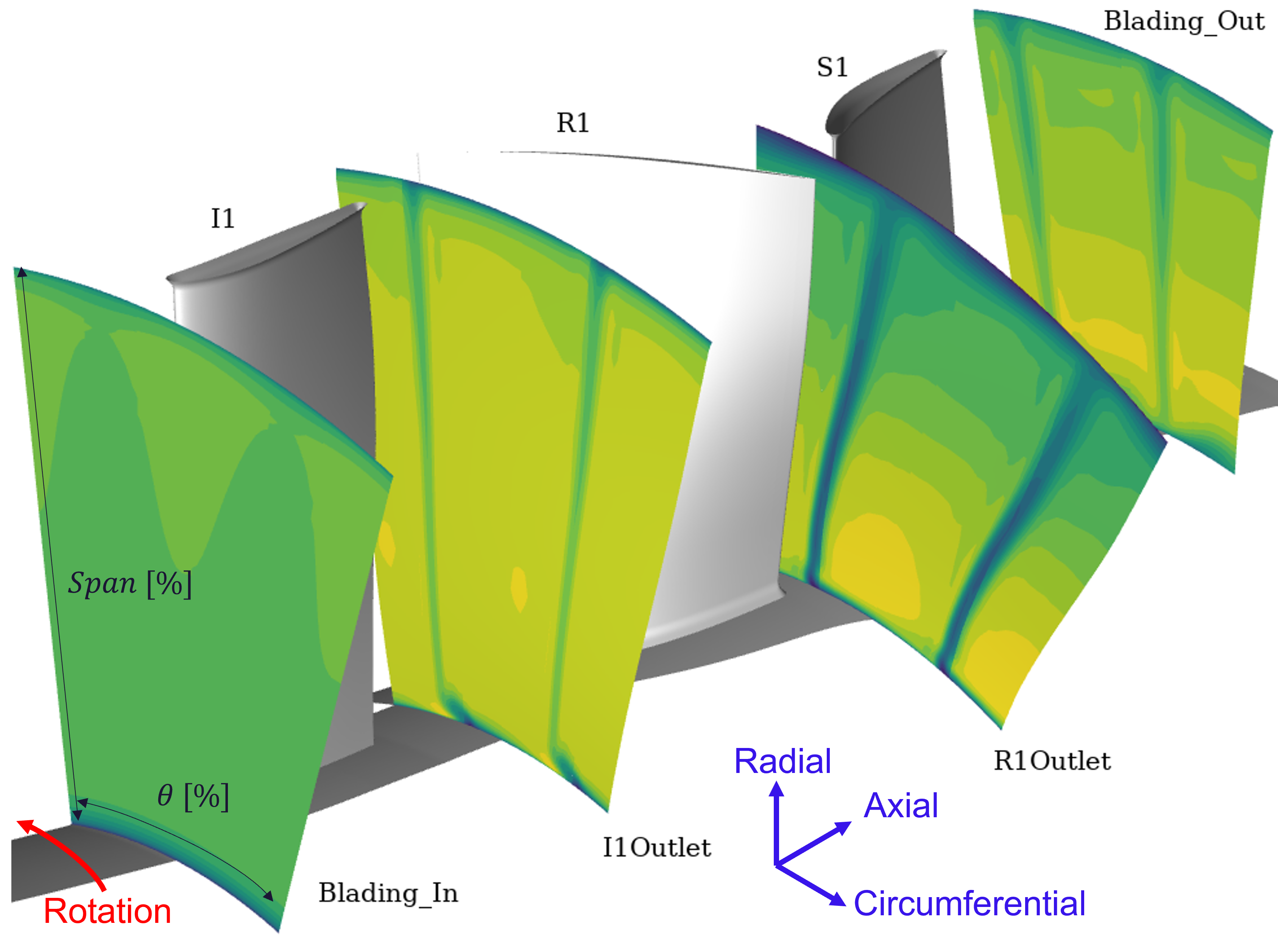}
       \caption{Overview of the CFD domain and locations selected for post-processing, with corresponding axial velocity contours}
       \label{fig:Domain}
\end{figure}

The CFD results for the configuration of interest are considered as ground-truth. The computational model consisted of a single passage model of the 1.5 stage, with mixing plane interfaces. The computational mesh was generated using Numeca AUTOGRID5, and the CFD solver used was Trace, with SST $K-\omega$ turbulence model. More details on the computational setup used is available in literature \cite{Bruni2022}. The tip clearance was varied for each row between 0.1\% and 2.0\% of blade height (i.e. span). The input variable space was sampled using latin-hypercubic sampling. This range is significantly larger than the tolerance typically specified for gas turbine applications, and was selected to demonstrate the robustness of the methodology to out-of-tolerance cases. The CFD results are processed to only extract six variables, namely, Total Pressure ($P_t$), Total Temperature ($T_t$), Axial Velocity ($V_x$), Tangential Velocity ($V_t$), Radial Velocity ($V_r$) and Density ($\rho$), at four inter-row locations, \textit{Blading\_In}, \textit{I1Outlet}, \textit{R1Outlet} and \textit{Blading\_Out}, as shown in Fig. \ref{fig:Domain}. 

\begin{figure}[!htbp]
    \centering
       \includegraphics[width=0.47\textwidth]{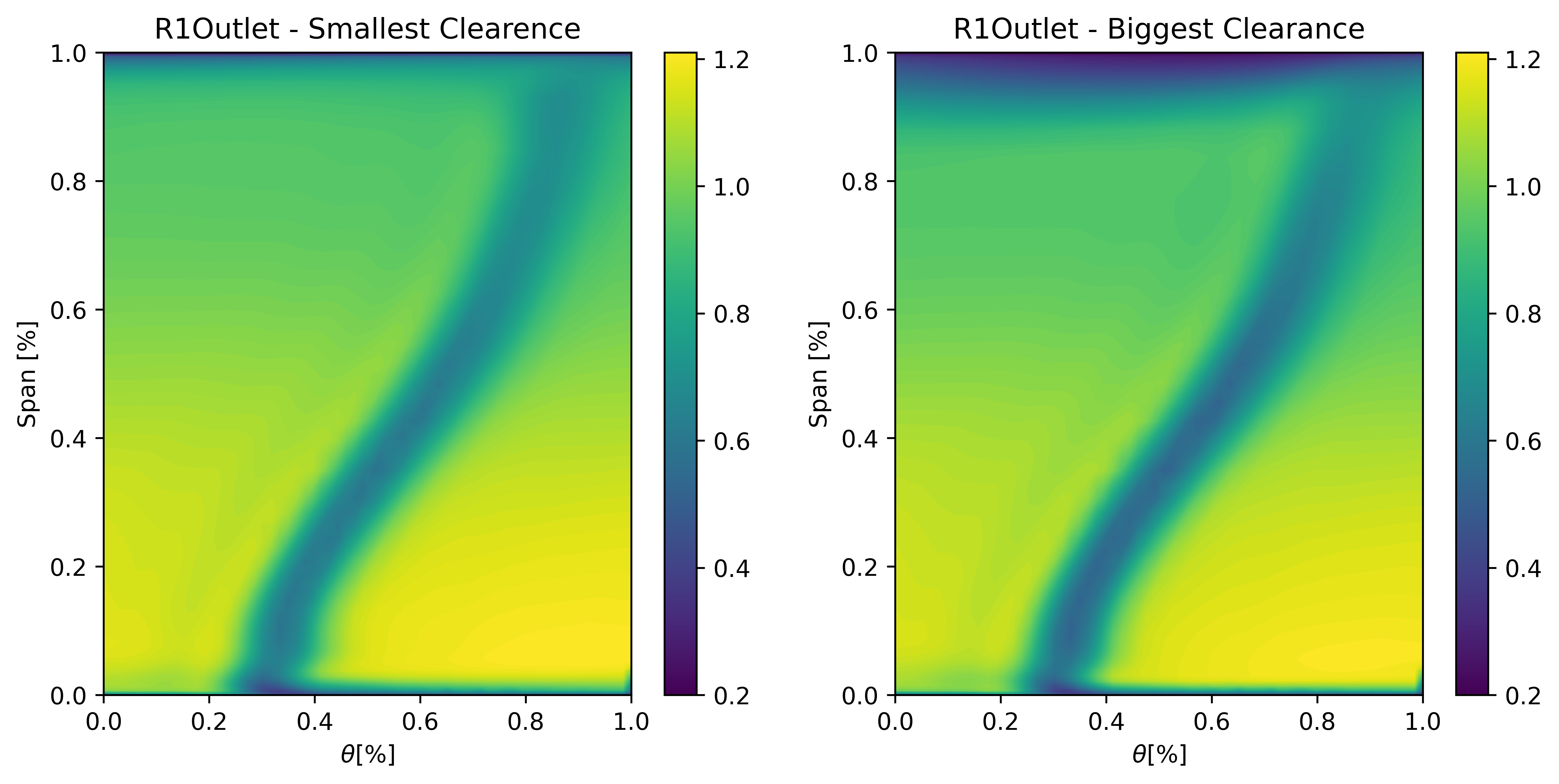}
       \caption{$V_x$ contours - smallest and largest clearance comparison}
       \label{fig:BiggestSmallest_Vx}
\end{figure}

\begin{figure}[!htbp]
    \centering
       \includegraphics[width=0.47\textwidth]{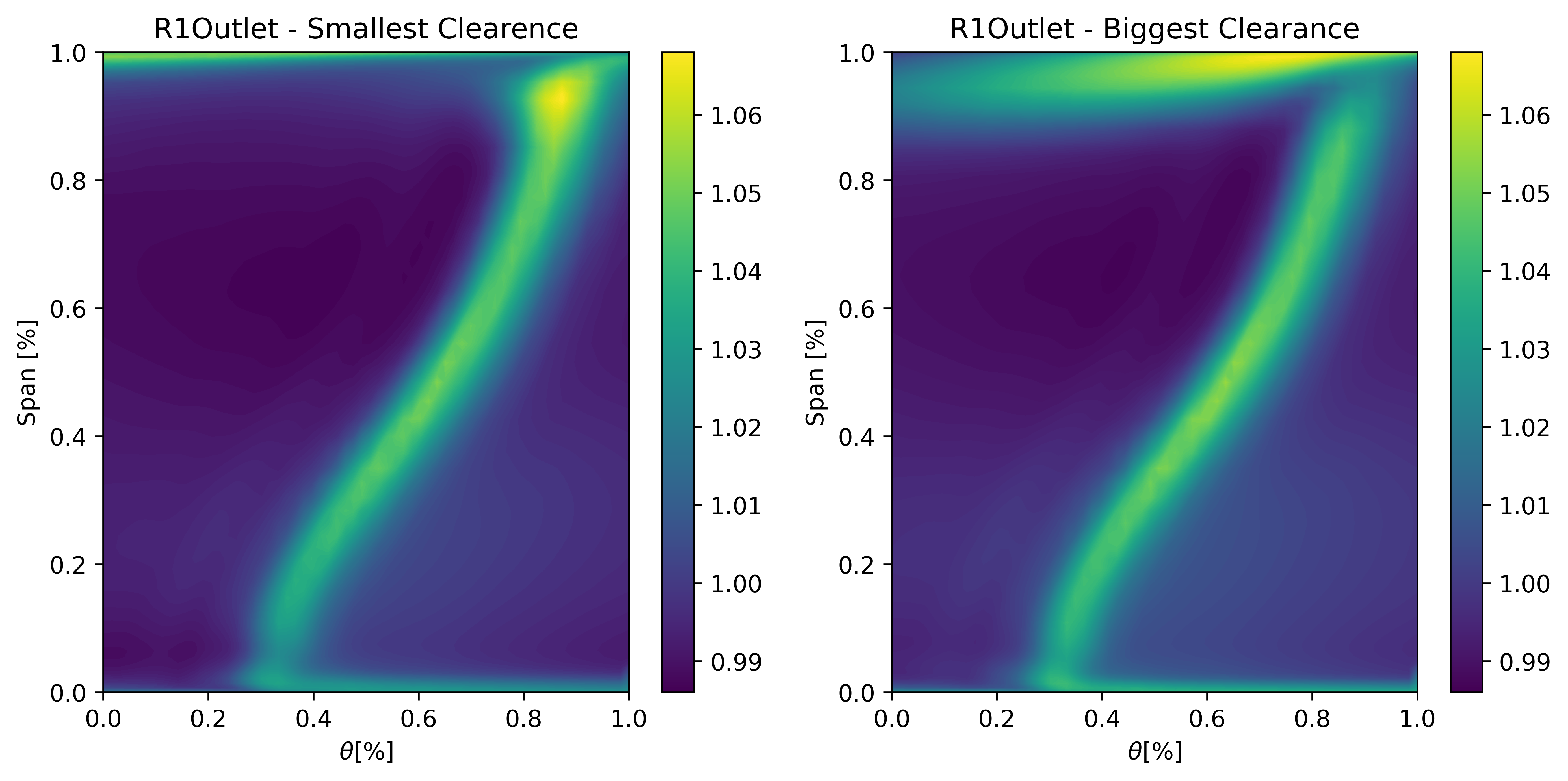}
       \caption{$T_0$ contours - smallest and largest clearance comparison}
       \label{fig:BiggestSmallest_Tt}
\end{figure}

These six values fully define the flow condition at each mesh node and can be used to calculate all the other flow variables available in the CFD solution. The CFD mesh for each simulation consists of more than one million nodes, yet the post-processed domain includes only 16 thousand nodes, while containing all the information required for the engineering assessment of interest. The data is stored in a tensor of shape $(4\times64\times64\times6)$, where each dimension represents the number of axial locations, tangential and radial nodes in the mesh, and number of variables respectively. While the current work considers only a single compressor geometry and operating point, the variation in clearance itself leads to a significant impact on the flow field. An overview of the impact of tip clearance variations on the flow field is presented by comparing the CFD results for the cases with the smallest and biggest clearances. All the values presented are non-dimensionalised based on the results from the nominal clearance baseline. The $Span [\%]$ y-axis represents the radial direction, going from the hub surface at the bottom,  to the casing surface at the top. The $\theta [\%]$ x-axis represents the circumferential direction, with the range between 0 and 1 consisting of one passage. Each row consists of a certain number of blades, and therefore the gas path can be divided in the circumferential direction in a number of passages equal to the blade count. The computational cost of the CFD analysis can be greatly reduced by assuming rotational periodicity and considering only a single passage, instead of the full annulus. Fig. \ref{fig:BiggestSmallest_Vx} shows the impact on $V_x$ and  Fig. \ref{fig:BiggestSmallest_Tt} on $T_0$ at \textit{R1Outlet}. 


\begin{figure}[!htbp]
    \centering
       \includegraphics[width=0.47\textwidth]{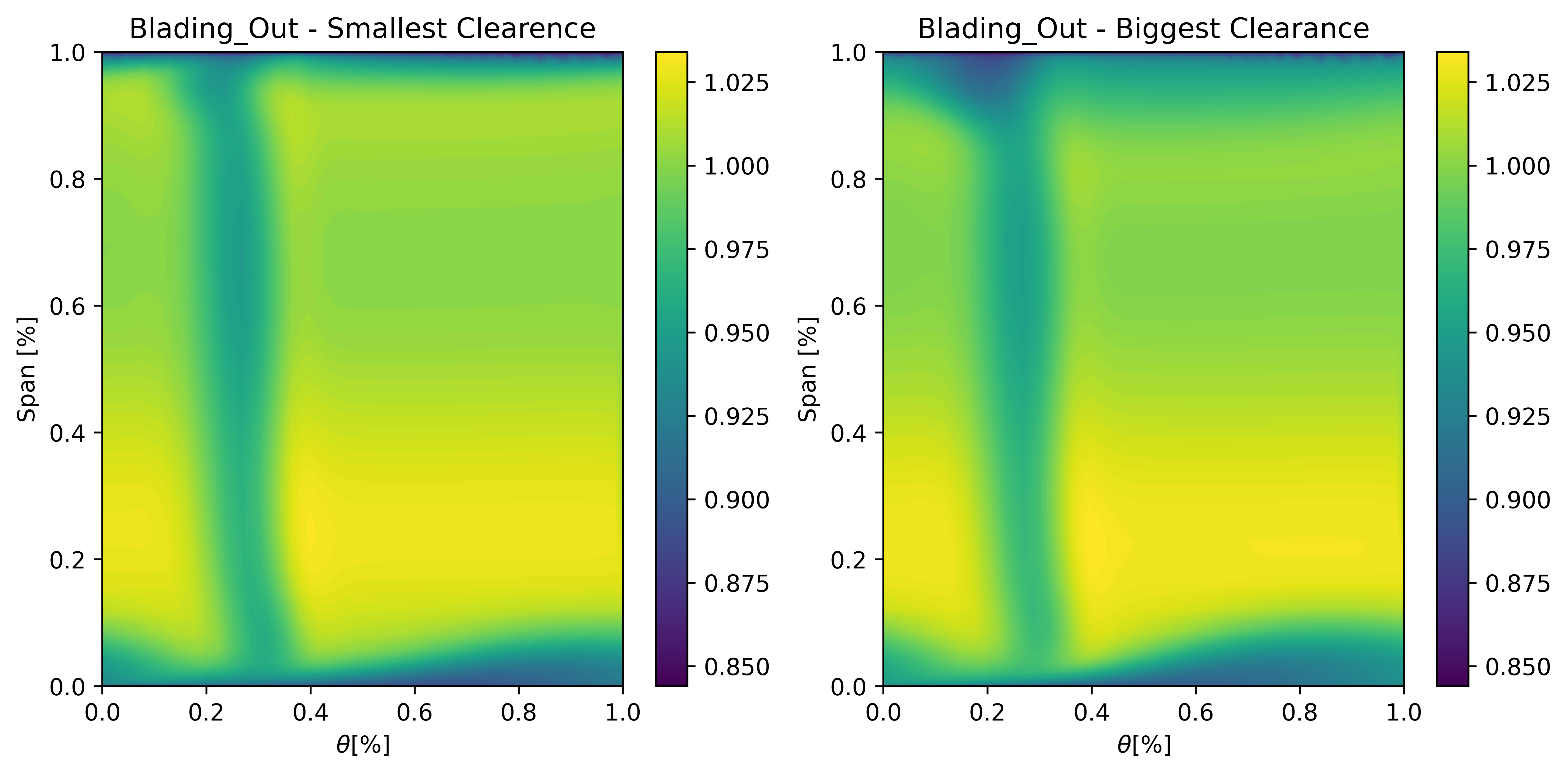}
       \caption{$P_0$ contours - smallest and largest clearance comparison}
       \label{fig:BiggestSmallest_Pt}
\end{figure}

The biggest clearance case presents lower axial velocity and higher total temperature in the casing region. This is due to the higher rotor tip leakage mass-flow upstream, which leads to a region of lower axial momentum, in turn then leading to an increase in local entropy generation due to mixing with the main passage flow. Fig. \ref{fig:BiggestSmallest_Pt} shows the impact on $P_0$ at \textit{Blading\_Out}. The case with the biggest clearance shows lower total pressure in the hub region due to the higher stator tip leakage losses upstream. Lower total pressure is found also in the casing due to the higher tip leakage losses in the upstream rotor, even if those have already partially mixed out with the main passage flow across the stator row. 

\begin{figure}[!htbp]
    \centering
       \includegraphics[width=0.47\textwidth]{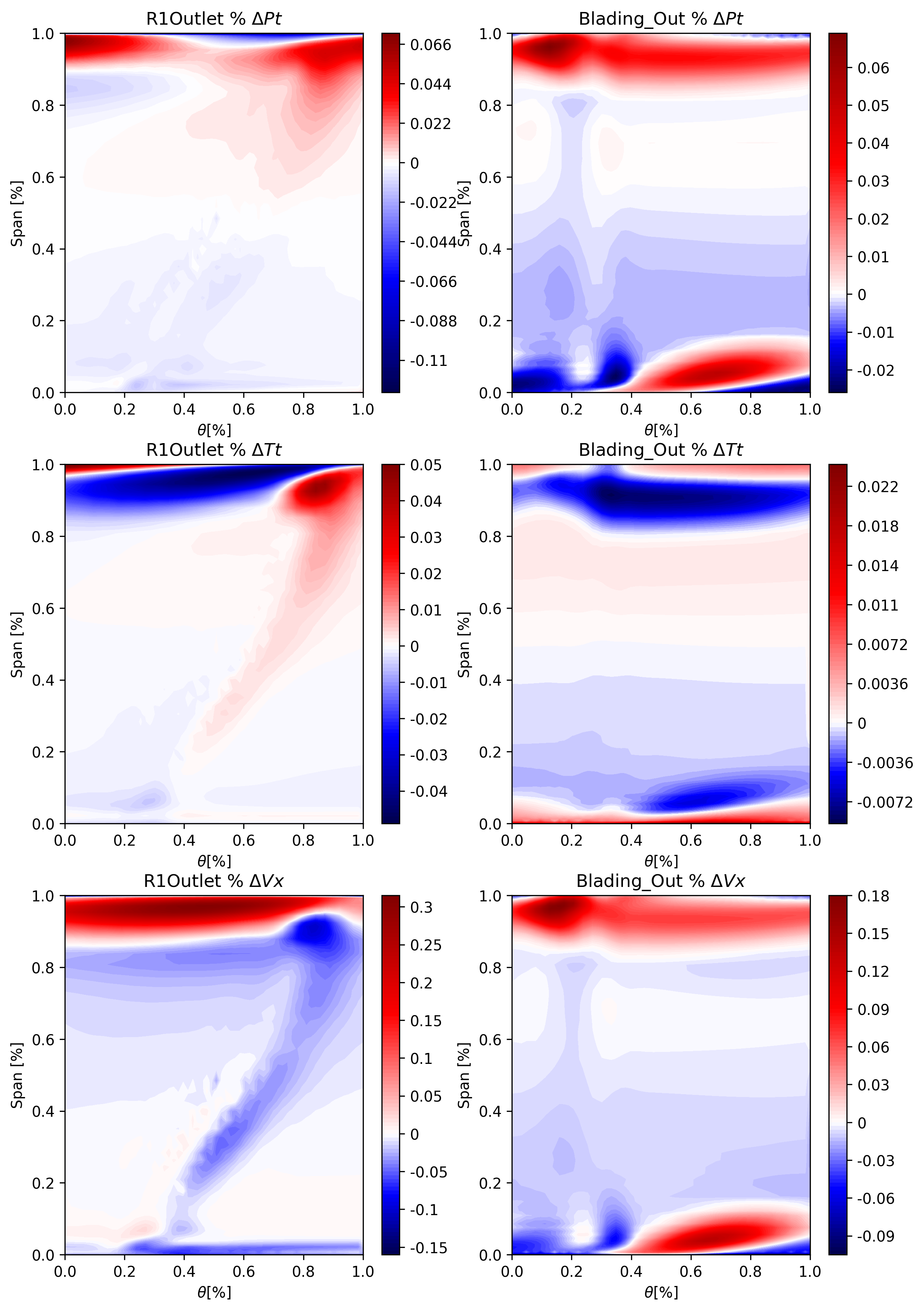}
       \caption{\% difference between smallest and largest clearance}
       \label{fig:BiggestSmallest_Delta}
\end{figure}

Fig. \ref{fig:BiggestSmallest_Delta} shows an overview of the percentage difference between the two cases for the different variables and the different locations. The greatest difference is found for $V_x$ in the casing region of \textit{R1Outlet}, with more than $0.3\%$.

\begin{figure}[!htbp]
    \centering
       \includegraphics[width=0.48\textwidth]{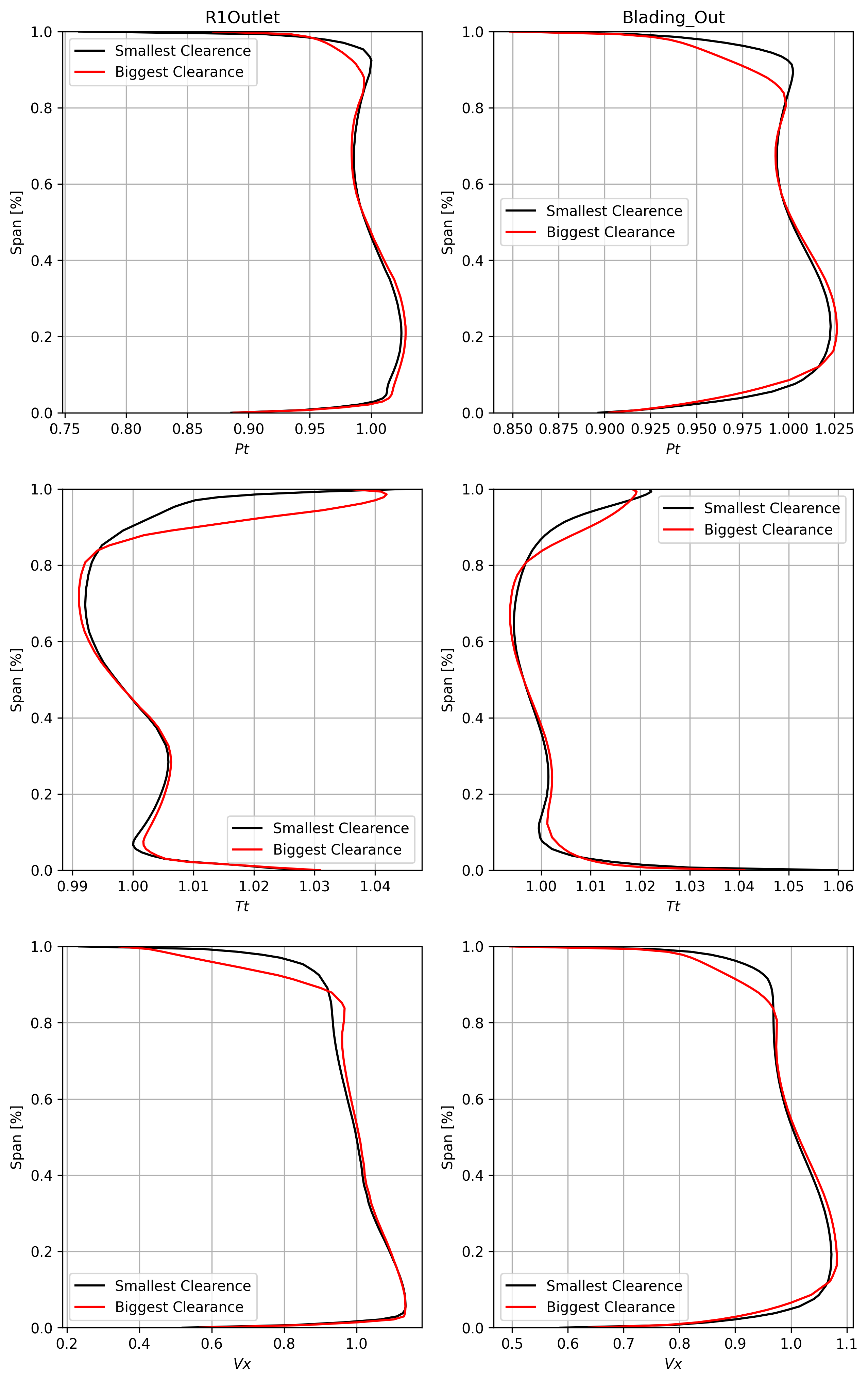}
       \caption{Radial profiles comparison: smallest vs largest clearance}
       \label{fig:BiggestSmaller_Radial}
\end{figure}

Usually, 2D contours are assessed only for very detailed analyses. For most engineering applications, it is more common to perform mass-flow averaging \cite{CumpstyAveraging} in the circumferential direction to obtain radial profiles of the variables of interest. The radial profiles in Fig. \ref{fig:BiggestSmaller_Radial} also show that significant differences are found between the two cases. The main variations are found near the hub and casing region, while the main passage flow is largely unaffected by variations in tip clearance. 

These radial profiles of each variable are then mass-flow averaged to obtain an 1D average, which are used to calculate the overall performance variables of interest, such as mass-flow $\dot{m}$ and polytropic efficiency $\eta_p$. The non-dimensional variation of those overall performance parameters between the smallest and largest clearance cases are $\Delta \dot{m} = 0.62\%$, and $\Delta \eta_p = 0.93 \%$. The range of the overall performance variables is provided for reference, to relate the errors in the machine learning predictions compared to typical variations for practical applications.
    
\newpage
\subsection{The C(NN)FD framework}

\begin{figure}[!htbp]
    \centering
       \includegraphics[width=0.48\textwidth]{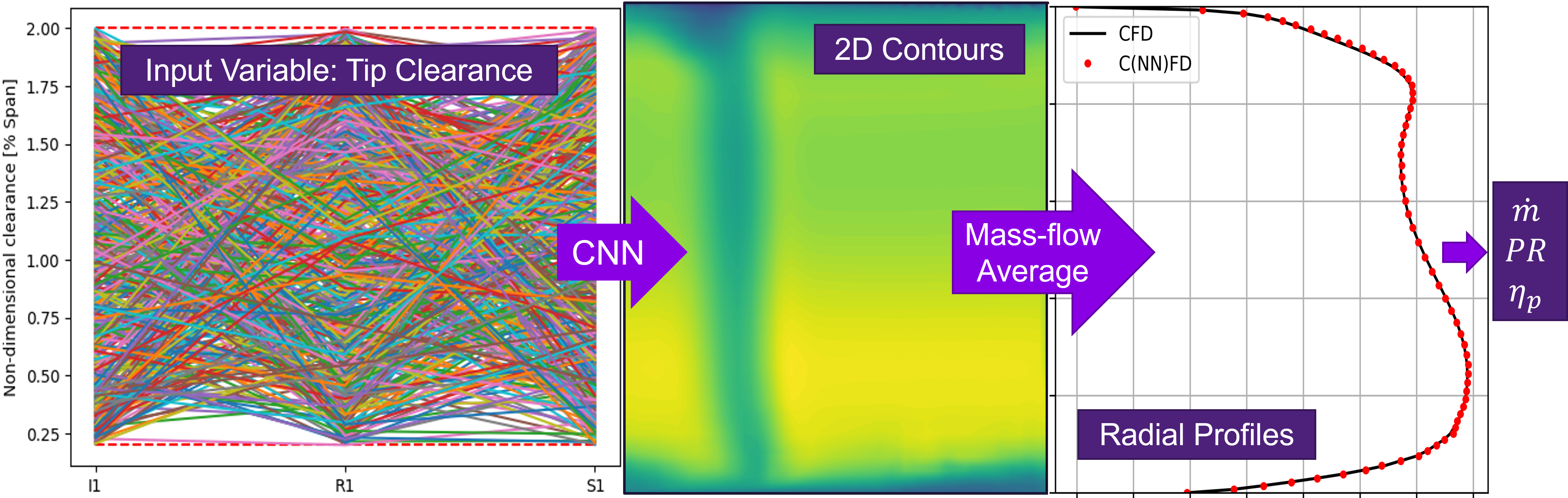}
       \caption{C(NN)FD - Framework overview}
       \label{fig:Framework}
\end{figure}

An overview of the proposed framework is summarised in Fig. \ref{fig:Framework}. The tip clearance values of a specific build and geometry design parameters are provided as an input to \textit{C(NN)FD}, which predicts the 2D contours for all the variables at the locations of interest. The outputs are then mass-flow averaged to obtain the relevant radial profiles and 1D averages, which are then used to calculate the overall performance using the relevant equations. The proposed framework aims at predicting 2D contours, radial profiles, 1D averages and overall performance concurrently, rather than directly targeting overall performance in a "black-box" approach. As the impact of tip clearances variations depends on the aerodynamic loading distribution of a given aerofoil, the methodology should be generalisable to different designs. \textit{C(NN)FD} therefore requires as input the geometries associated with each clearance value. This is achieved by considering a series of design parameters to describe the geometry of each blade, such as stagger angle, camber angle, maximum thickness, etc. In this paper, only a single compressor design is considered, and therefore the geometry parameters are fixed. Future work will consider the effect of geometrical variations and different designs.

\begin{figure}[!htbp]
    \centering
       \includegraphics[width=0.49\textwidth]{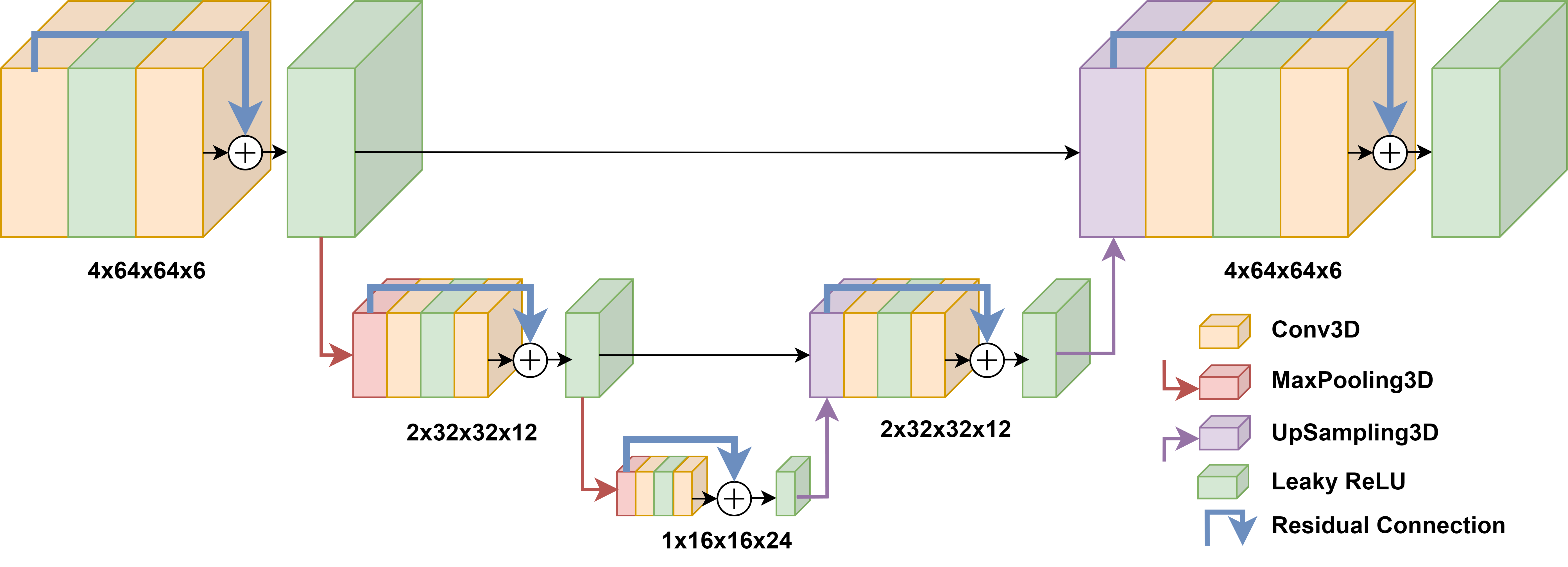}
       \caption{C(NN)FD architecture overview}
       \label{fig:TII-23-3364_architecture}
\end{figure}

The generated dataset consists of 500 CFD solutions, each of which was run using automated and parallelized processes, taking 15 minutes on 24 CPUs for each case. The dataset was split into 70\% training data, 20\% validation data and 10\% hold-out data. The validation set was used to evaluate the performance of the model and to perform hyper-parameter tuning, while the hold-out set was set aside and was considered only for a final assessment presented in this paper. The C(NN)FD architecture illustrated in Fig. \ref{fig:TII-23-3364_architecture} is a 3D variant of the U-Net \cite{Thuerey2020}, modified to include residual connections for each convolutional block. The advantage of this architecture is that the encoding section, followed by the decoding section in conjunction with the skip connections, allow to predict both low level and high level features in the flow field. The use of residual connections improves convergence and allows for this architecture type to be scalable to larger models. Batch Normalization is used before each convolutional layer, the activation function used is \textit{Leaky Relu} with a slope of 0.2 for all layers, the weights are initialized using \textit{He normal} and the network is trained using the \textit{Adam} optimiser, with \textit{Mean-Squared-Error} as loss function, a batch size of 40, and a learning rate of 0.002. The selected architecture was found to be the best performing in comparison to a series of fully connected and convolutional neural networks that were considered during development. 

While U-Nets are a reasonable compromise between model complexity and accuracy of the predictions, future work could potentially consider more advanced methods such as recurrent neural networks, generative adversarial network or graph convolutional neural networks. The first input of the network is a tensor with the tip clearance values of size $(3\times1)$, representing respectively the number of rows, and the tip clearance values. The second input is a tensor with the blade geometry design parameters of size $(3\times8\times5)$, which represent the number of rows, number of radial sections, and number of design variables. The two tensors are then added and duplicated to match the dimensions required by the U-Net input of $(4\times64\times64\times6)$. The output of the network is also a tensor of size $(4\times64\times64\times6)$ describing the whole flow field. The width of the network here is limited to two down-sampling steps due to the stride of $(2,2,2)$ selected for the max-pooling operation, which leads to a bottle-neck section with tensors of shape $(1\times16\times16\times24)$. This is because the number of axial locations for this test case is only four. Future applications of multi-stage axial compressors will require a higher number of axial locations, potentially 24 or more depending on the number of stages. The overall size of the architecture is of only $\sim50k$ trainable parameters, which results in training times in the order of magnitude of minutes, making the architecture scalable to industrial applications. The execution time for inference is less than 1 second, making the predictions of the deployed model effectively real-time for the application of interest. 

\section{Results}

\begin{figure}[!htbp]
    \centering
       \includegraphics[width=0.49\textwidth]{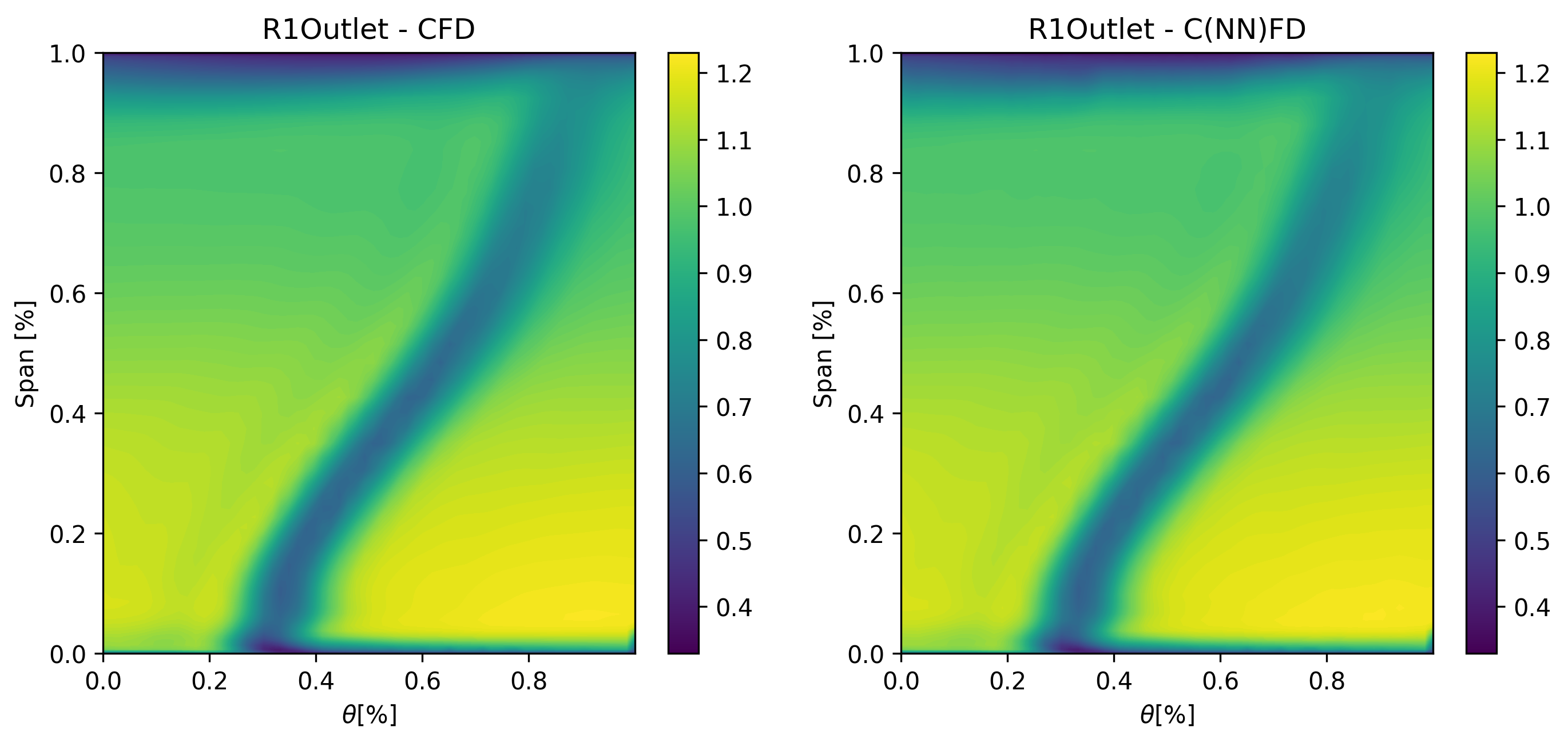}
       \caption{$V_x$ contours - CFD ground truth and ML prediction comparison}
       \label{fig:2DWorst_Vx}
\end{figure}

\begin{figure}[!htbp]
    \centering
       \includegraphics[width=0.49\textwidth]{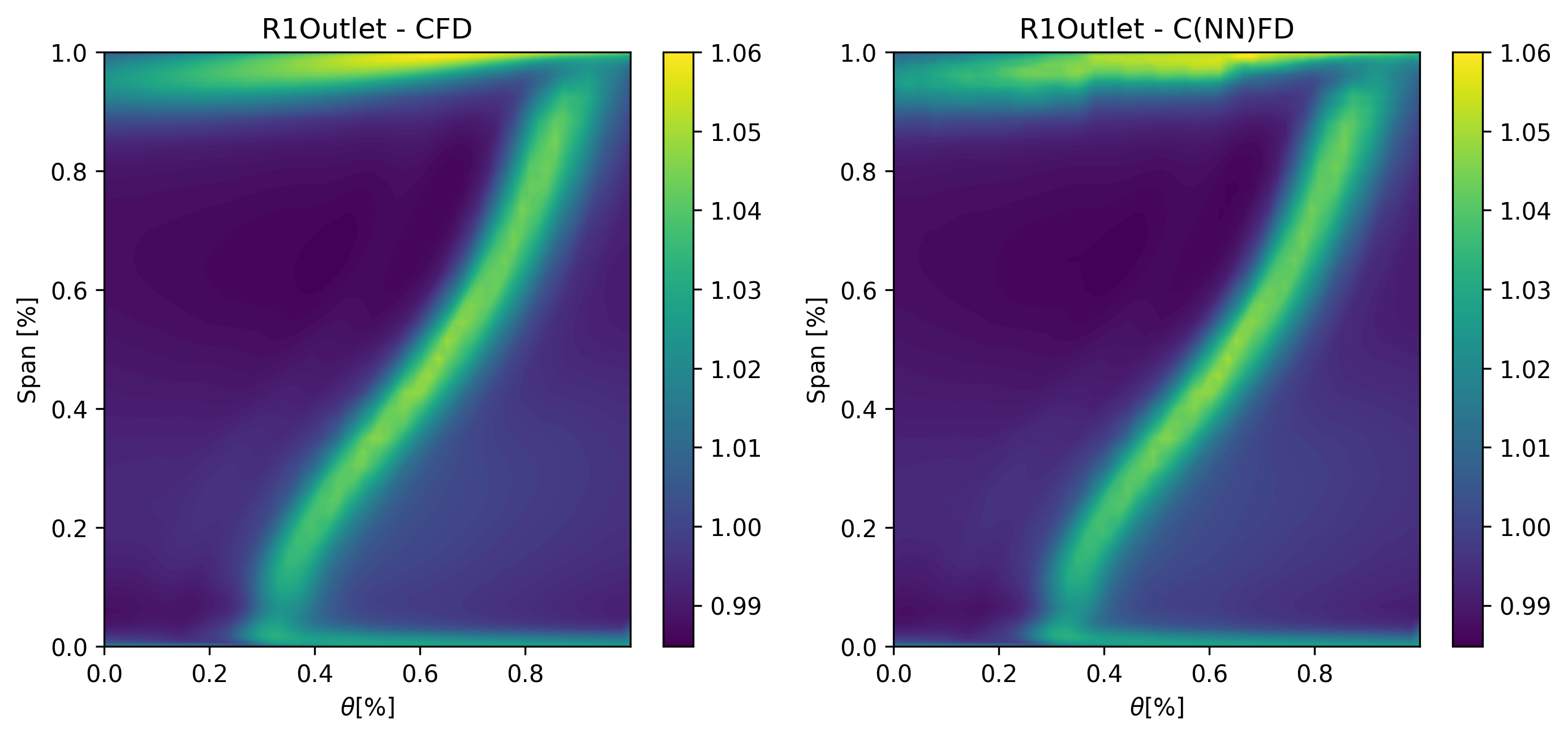}
       \caption{$T_t$ contours - CFD ground truth and ML prediction comparison}
       \label{fig:2DWorst_Tt}
\end{figure}

\begin{figure}[!htbp]
    \centering
       \includegraphics[width=0.49\textwidth]{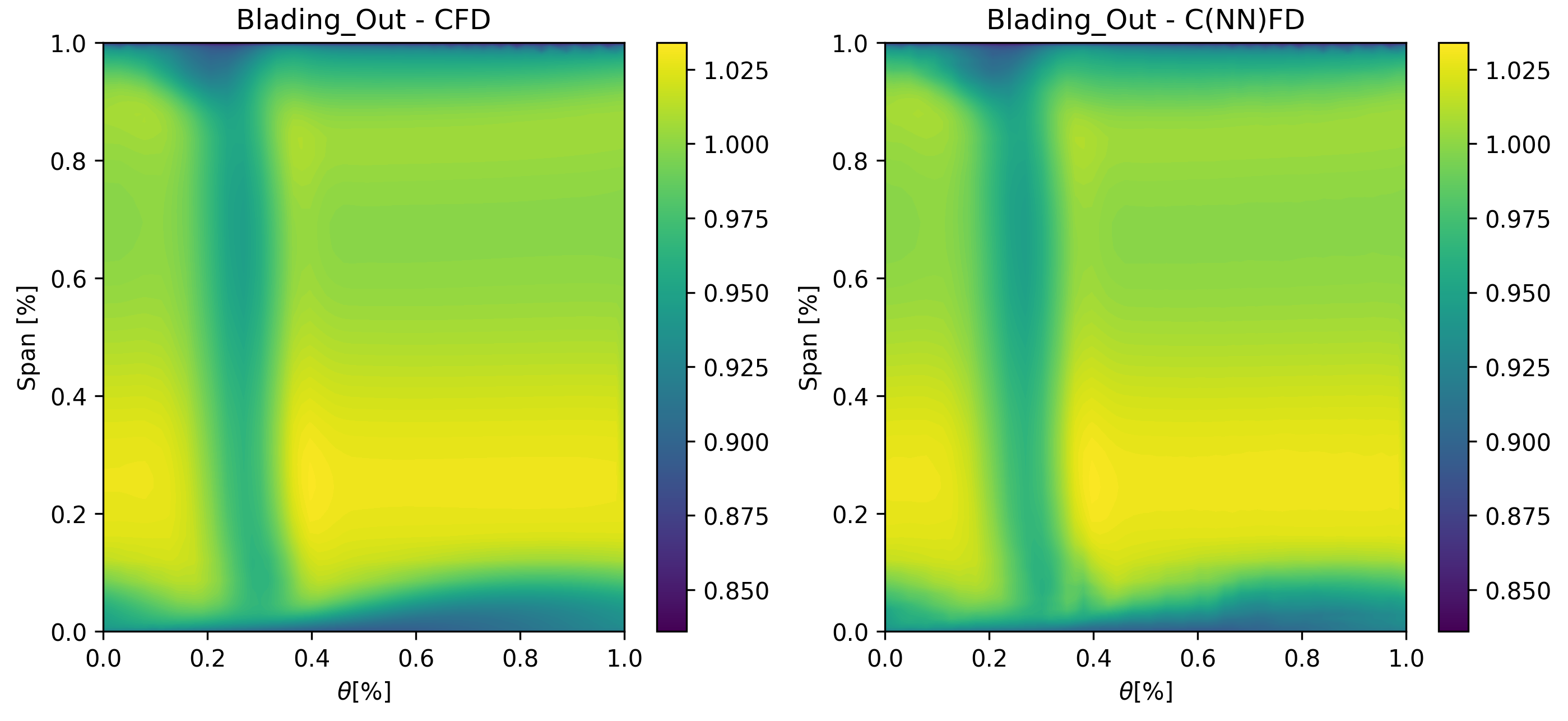}
       \caption{$P_t$ contours - CFD ground truth and ML prediction comparison}
       \label{fig:2DWorst_Pt}
\end{figure}

The comparison between the predicted flow field from \textit{C(NN)FD} and the CFD ground truth is presented for the worst performing case, where the prediction error was found to be the greatest in the holdout set. An overview of the differences in the flow field is presented by comparing the ground truth on the left, to the predictions on the right. The same locations and variables discussed in the comparison between the smallest and the largest clearance are presented. Fig. \ref{fig:2DWorst_Vx} shows the comparison for Axial Velocity and Fig. \ref{fig:2DWorst_Tt} for Total Temperature at \textit{R1Outlet}. Fig. \ref{fig:2DWorst_Pt} shows the comparison for Total Pressure at \textit{Blading\_Outlet}. Overall, excellent agreement is found, with the main flow features being reproduced correctly by the predictions. Only negligible differences can be found in the regions of high gradients, where the predictions are not as smooth as the ground truth. This is visually noticeable only for Fig. \ref{fig:2DWorst_Tt} near the casing where the errors are in the order of magnitude of less than $0.01 \%$, which corresponds to less than $0.1 ^\circ C$ and are well within the numerical uncertainty of the CFD solver. 


\begin{figure}[!htbp]
    \centering
       \includegraphics[width=0.45\textwidth]{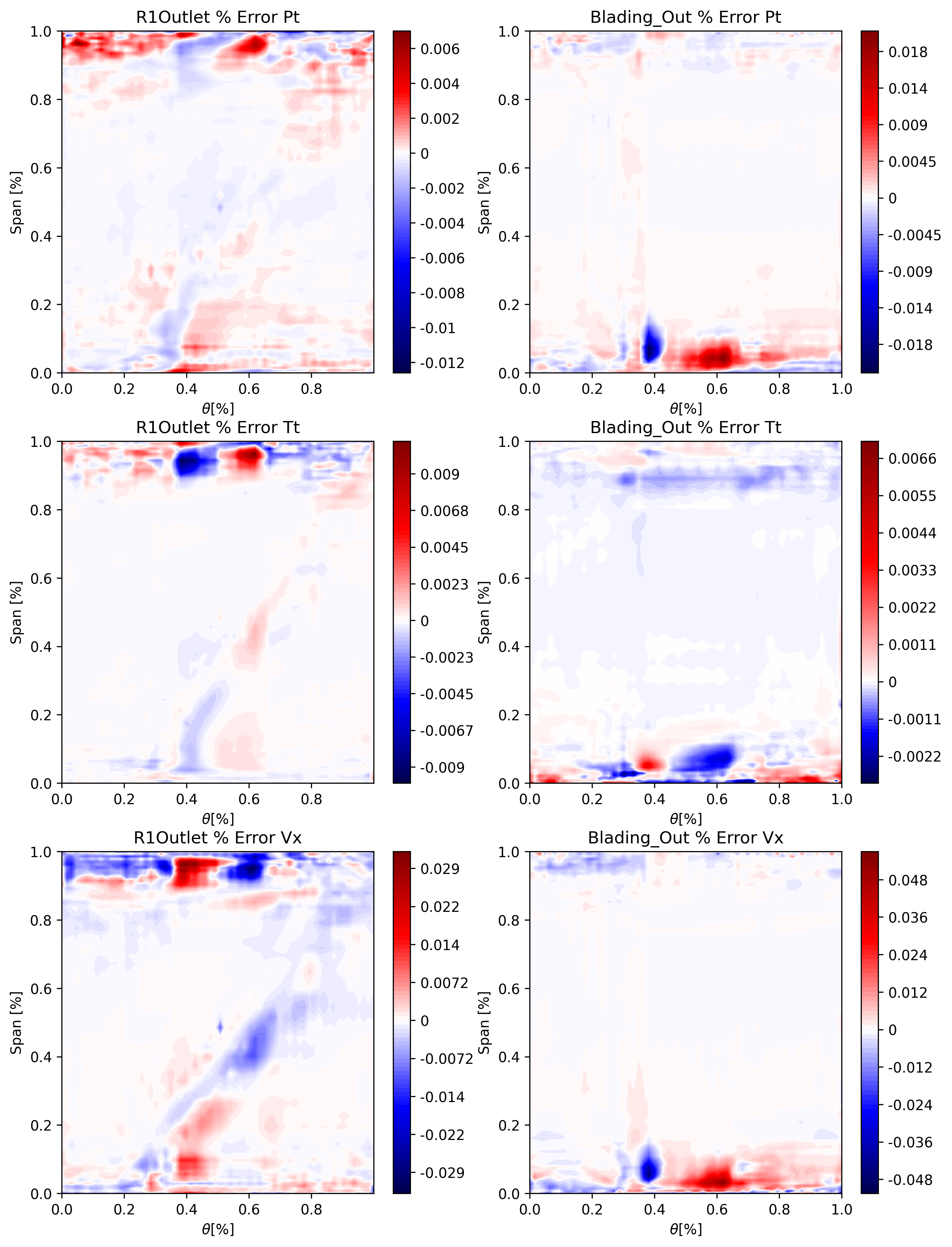}
       \caption{2D contours \% error between ground truth and predicted values}
       \label{fig:2DWorst_Error}
\end{figure}

\begin{figure}[!htbp]
    \centering
       \includegraphics[width=0.45\textwidth]{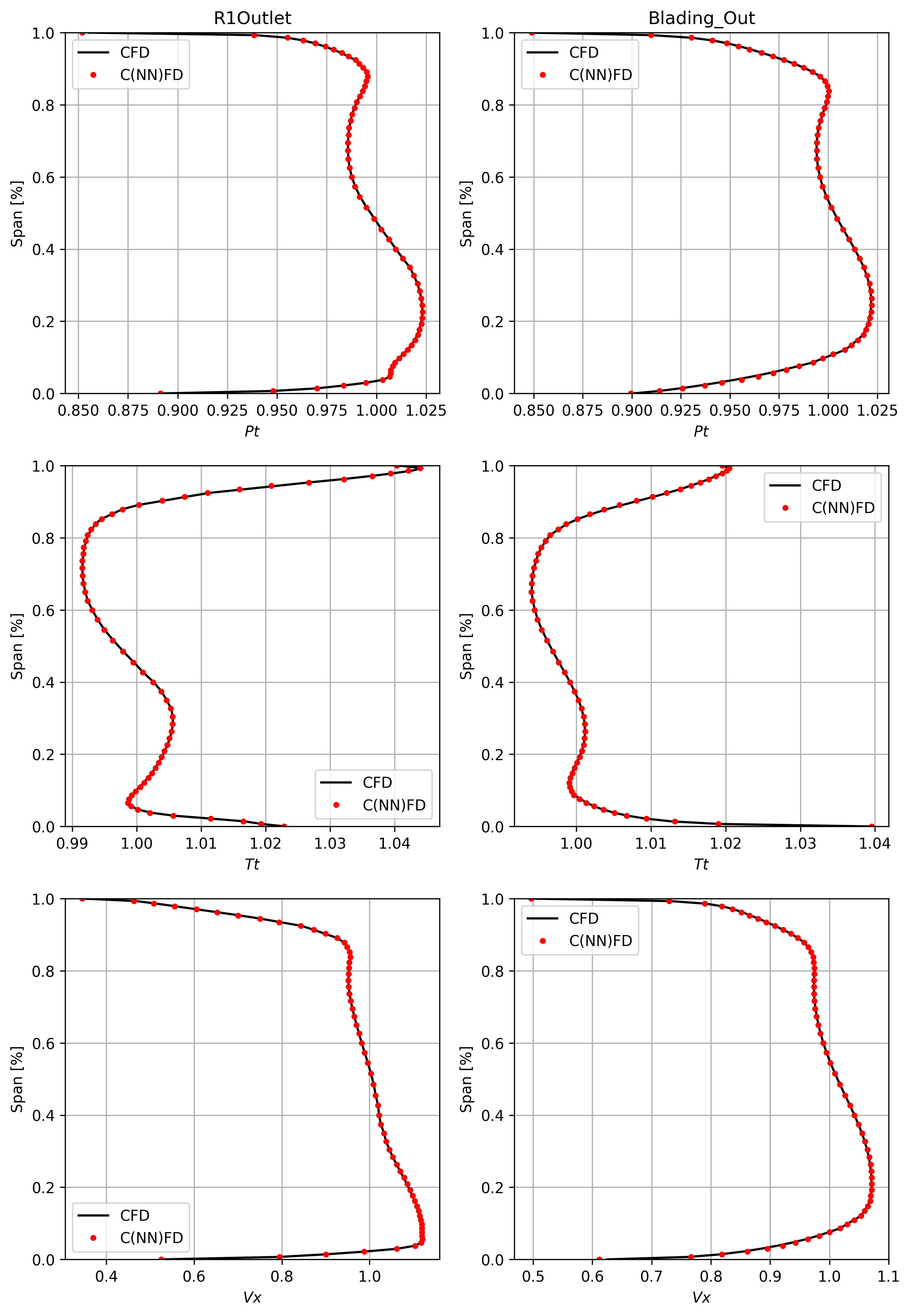}
       \caption{Radial profiles comparison: ground truth vs predicted values}
       \label{fig:Worst_Radial}
\end{figure}

Fig. \ref{fig:2DWorst_Error} shows an overview of the error between predictions and ground truth, with a peak value of $0.048 \%$ in the axial velocity at \textit{Blading\_Out}. This is less than one order of magnitude smaller compared to the range shown in Fig. \ref{fig:BiggestSmallest_Delta} for the comparison between smallest and largest clearance. Therefore, the actual error in the contours is negligible when compared to the range in the dataset. Moreover, the errors in the 2D contours predictions are very localised to regions with steep gradients and to only few nodes in the mesh. When performing mass-flow averaging to obtain the radial profiles, the discrepancies between ground truth and predictions are even less significant. Fig. \ref{fig:Worst_Radial} shows how the radial profiles for \textit{C(NN)FD} and CFD are effectively overlapping. It can be noticed how even the end-wall regions characterized by steep gradients are well resolved by \textit{C(NN)FD}. 

\newpage

\begin{figure}[!!htbp]
    \centering
       \includegraphics[width=0.45\textwidth]{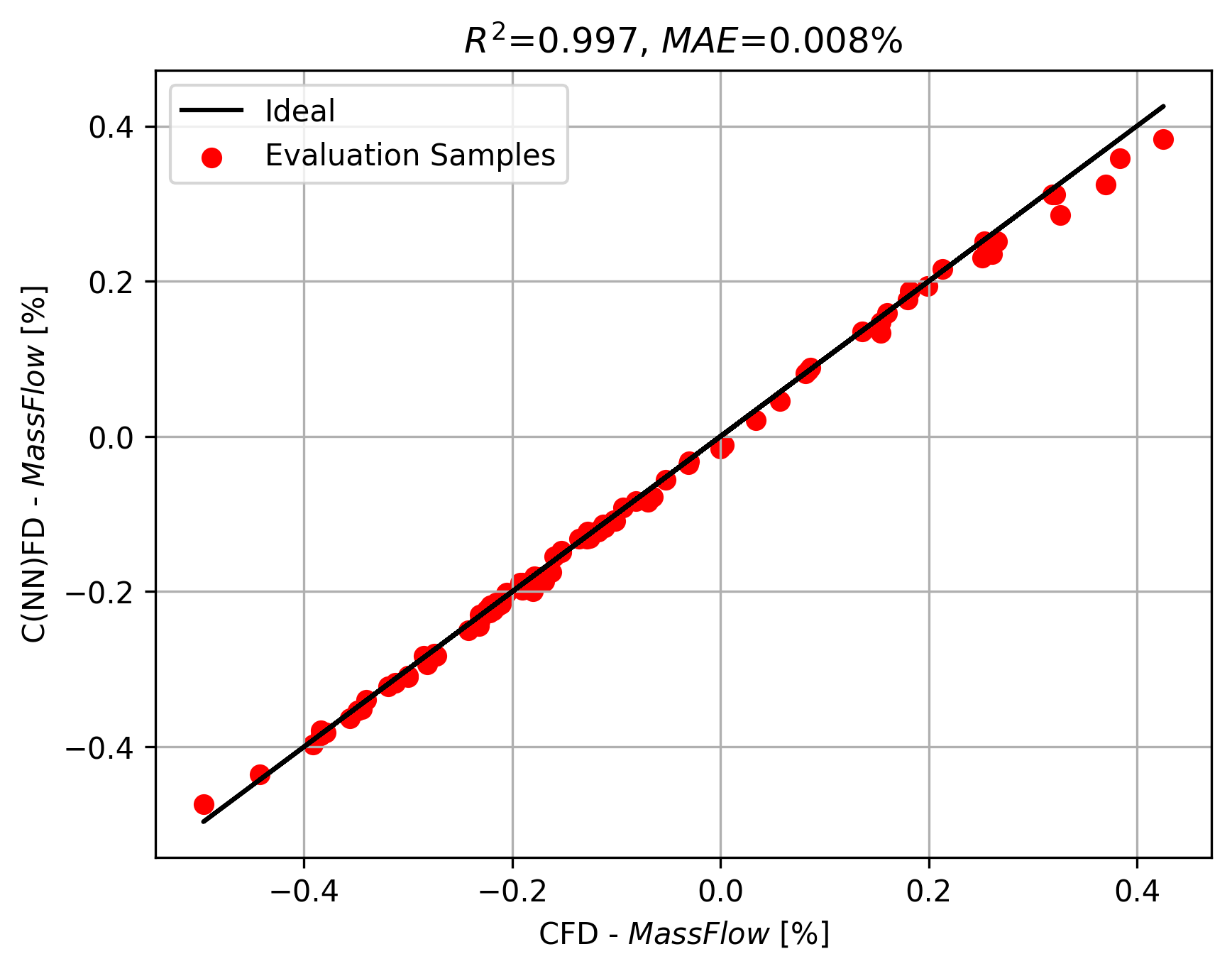}
       \caption{Mass-flow predictions by (C(NN)FD) and ground-truth (CFD)}
       \label{fig:Overall_Mass}
\end{figure}

\begin{figure}[!!htbp]
    \centering
       \includegraphics[width=0.45\textwidth]{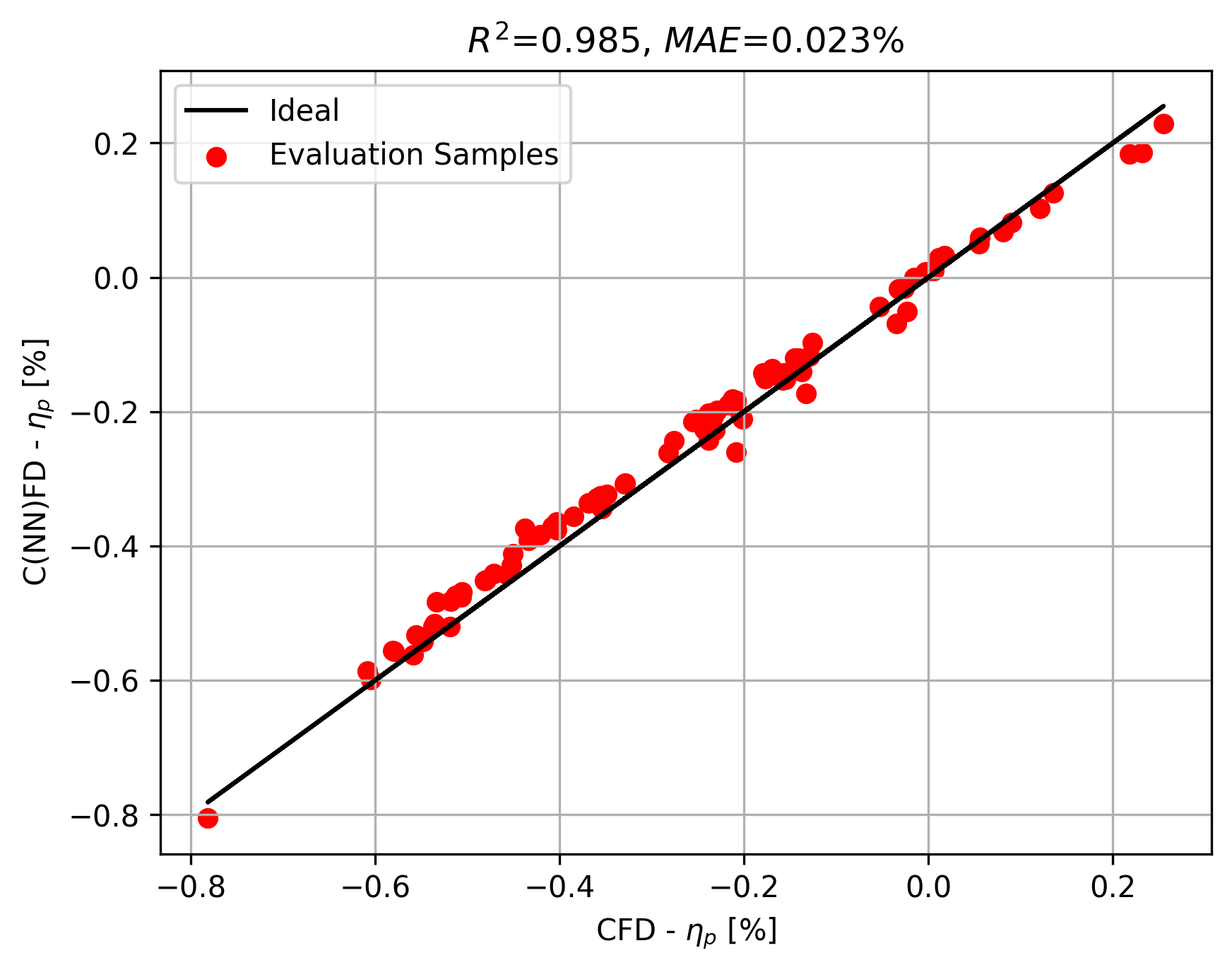}
       \caption{$\eta_p$ predictions (C(NN)FD) and ground-truth (CFD)}
       \label{fig:Overall_EtaP}
\end{figure}

The radial profiles are then further mass-flow averaged to obtain the relevant 1D averages, which are used to calculate the overall performance variables of interest. An overview is presented in Table \ref{table:OverallPerformance}, Fig. \ref{fig:Overall_Mass} and Fig. \ref{fig:Overall_EtaP}, showing the comparison between predicted values and ground truth respectively for mass-flow, pressure-ratio and polytropic efficiency. An excellent agreement is found for all the overall performance parameters, with all the variables  showing a coefficient of determination $R^2$ close to 1, suggesting that the proposed model can accurately describe most of the variance found in the dataset. Likewise, the Mean-Absolute-Error for each variable is smaller than the known uncertanties of the results of the CFD used for the ground truth, and significantly smaller than range of the dataset.

\begin{table}[!htbp]
    \centering    
    \caption{Overall performance comparison}
    \begin{tabular}{c|ccc}
     
     Var       & $R^2$ & $MAE \%$   & $\Delta dataset \%$   \\ \hline
     $\dot{m}$ & 0.997 & 0.008          &  0.62              \\      $\eta_p$  & 0.985 & 0.024          &  0.93              \\ 
    \end{tabular}
\label{table:OverallPerformance}
\end{table}


The \textit{C(NN)FD} predictions were found to be accurate even for the case with the lowest efficiency in the hold-out set, showing the capability of modelling challenging aerodynamic conditions typical of large clearances, as long as being represented in the training data available. Future work will extend the methodology to consider a multi-stage compressor application with additional sources of manufacturing and build variations such as surface roughness and geometrical variations. This will lead to a more high-dimensional problem, which might potentially require more complex architectures, or greater availability of training data.

\section{Conclusion}

This paper demonstrated the development of a novel deep learning framework for real-time predictions of the impact of manufacturing and build variations on the flow field and overall performance of axial compressors. The proposed \textit{C(NN)FD} architecture was shown to achieve  real-time accuracy comparable to the CFD benchmark. Predicting the flow field and using it to calculate the corresponding overall performance renders the methodology generalisable, while filtering only relevant parts of the CFD solution, renders the methodology scalable to industrial applications. The initial application considered the impact of tip clearance variations on a 1.5 stages axial compressor. Future work will extend the methodology to consider a multi-stage compressor application, as well as including other sources of manufacturing and build variations such as surface roughness and geometrical variations. 

\bibliographystyle{IEEEtran}
\bibliography{TII-23-3364}
\begin{IEEEbiographynophoto}{Giuseppe Bruni}
    is currently employed as a Principal Aerodynamicist by Siemens Energy Industrial Turbomachinery Ltd, Lincoln, UK, where he has been working since 2016. He received the BSc and MSc degrees in Mechanical Engineering from University of Padova, Italy, respectively in 2014 and 2016. He received the MSc in Gas Turbine Technology from Cranfield University, UK in 2016, and is currently working towards a PhD at the University of Lincoln, UK, on the topic of Machine learning modelling of manufacturing variations on compressor aerodynamic performance. His research interest include the aerodynamic, aero-mechanical analysis and design of axial compressors, as well as development and applications of optimisation and machine learning methods. He is a Chartered Engineer and member of the IMechE.
\end{IEEEbiographynophoto}
\begin{IEEEbiographynophoto}{Dr. Sepehr Maleki}
    is a Senior Lecturer in AI and Industrial Digitalisation at the School of Engineering of the University of Lincoln, where he also serves as the Chair of the Digitalisation Research Network. His research expertise includes modelling intricate industrial systems and developing multi-agent systems that enhance productivity and efficiency, with a specific focus on machine learning-based techniques. Sepehr obtained his MSc in Wireless Communication in 2011 and his PhD in 2015 from the University of Southampton. He served as a Research Fellow in the School of Engineering at the University of Lincoln from 2014 to 2019. He is currently a Member of the IET and a Fellow of the HEA. 
\end{IEEEbiographynophoto}
\begin{IEEEbiographynophoto}{Prof. Senthil K. Krishnababu}
    is currently employed by Siemens Energy Industrial Turbomachinery Ltd, UK as Manager of Test \& Validation. He was formerly a Group Leader for compressor aerodynamics \& mechanical integrity and Advisory Key Expert for aeromechanics. He is a visiting professor of Applied Machine Learning at University of Lincoln. He received his BEng in Mechanical Engineering in 1999 and MSc in Aerospace Engineering in 2002, from Indian Institute of Technology Madras, India. He was awarded Ph.D. in Engineering by Cambridge University in 2006. He was a post-doctoral researcher at Vibration UTC, Imperial college, London. He is a Fellow of IMechE and HEA
\end{IEEEbiographynophoto}
\end{document}